\documentclass[conference]{IEEEtran}
\IEEEoverridecommandlockouts

\usepackage{graphicx}
\usepackage{amsmath, bm}
\usepackage{amsfonts}
\usepackage{amssymb}
\usepackage{longtable}

\usepackage{import}
\usepackage{color}
\usepackage{amsthm}
\usepackage{cite}
\usepackage{scalerel,amssymb}

\usepackage{amssymb}
\usepackage{algorithmic}
\usepackage{algorithm}
\usepackage{rotating}
\newcommand{\floor}[1]{\left\lfloor #1 \right\rfloor}

\usepackage{lipsum}
\usepackage{multicol}
\usepackage{multirow}
\usepackage{url}
\usepackage{enumitem}
\usepackage{arydshln}
\usepackage{comment}

\usepackage{hyperref}

\def\BibTeX{{\rm B\kern-.05em{\sc i\kern-.025em b}\kern-.08em
    T\kern-.1667em\lower.7ex\hbox{E}\kern-.125emX}}

\newtheorem{myremark}{Remark}

\DeclareMathOperator*{\argmin}{arg\,min}

\begin{document}

\title{Use of Deterministic Transforms to Design \\ Weight Matrices of a Neural Network}


\author{\IEEEauthorblockN{Pol Grau Jurado, Xinyue Liang, Alireza M. Javid, and Saikat Chatterjee}
		\IEEEauthorblockA{\textit{School of Electrical Engineering and Computer Science} \\
			\textit{KTH Royal Institute of Technology, Sweden}\\
			\{polgj, xinyuel, almj, sach\}@kth.se}
	}


\maketitle
\IEEEtitleabstractindextext{
\begin{abstract}
Self size-estimating feedforward network (SSFN) is a feedforward multilayer network. For the existing SSFN, a part of each weight matrix is trained using a layer-wise convex optimization approach (a supervised training), while the other part is chosen as a random matrix instance (an unsupervised training). In this article, the use of deterministic transforms instead of random matrix instances for the SSFN weight matrices is explored. The use of deterministic transforms provides a reduction in computational complexity. The use of several deterministic transforms is investigated, such as discrete cosine transform, Hadamard transform, Hartley transform, and wavelet transforms. The choice of a deterministic transform among a set of transforms is made in an unsupervised manner. To this end, two methods based on features' statistical parameters are developed. The proposed methods help to design a neural net where deterministic transforms can vary across its layers' weight matrices. The effectiveness of the proposed approach vis-a-vis the SSFN is illustrated for object classification tasks using several benchmark datasets.
\end{abstract}

\begin{IEEEkeywords}
Multilayer neural network, deterministic transforms, weight matrices.
\end{IEEEkeywords}}

\IEEEdisplaynontitleabstractindextext

%
\IEEEpeerreviewmaketitle

\section{Introduction}\label{sec:Introduction} 

Over the past decade, the field of machine learning is enriched with appropriately trained neural network architectures such as deep neural networks (DNNs) \cite{DNN_2013} and convolutional neural networks (CNNs) \cite{CNN_2012}, outperforming the classical methods in different classification and regression problems \cite{Russakovsky2015,DodgeK17b}. However, this impressive progress has come with a cost--it has made machine learning strictly reliant on extensive computational resources such as parallel computations using graphical processing units (GPUs). 
    
There exist two main classes of algorithms that try to address the high computational complexity requirements of modern neural networks. The first class of algorithms tries to preserve the state-of-the-art performance of famous architectures, such as ResNet \cite{he2015residual} and AlexNet \cite{CNN_2012}, while reducing the size of the network as much as possible. EfficientNet \cite{EfficientNet_2019}, SqueezeNet \cite{SqueezeNet_2016}, and MobileNet \cite{MobileNet_2017} are examples of this kind. However, training of the above architecture is still quite expensive due to the use of stochastic gradient descent and backpropagation \cite{ruder2017overview}. The second class of algorithms tries to resolve this issue by using a gradient-free training approach. To this end, one popular technique is that some of the weight matrices of the network are set to instances of random matrices and only the rest of the weight matrices are updated during training. This leads to a convex relaxation of the training cost and eliminates the need for error backpropagation throughout the layers. Extreme learning machine (ELM) \cite{elm_Huang2012}, random vector functional link (RVFL) and its variants \cite{RVFL_1992,DRVFL_2019}, progressive learning network (PLN) \cite{PLN_Saikat}, and self size-estimating feedforward network (SSFN) \cite{SSFN_Saikat} are examples of this class that are shown to provide competitive performance with very low computational requirements in various applications. 

\textbf{Article contribution:} it is investigated the prospect of using deterministic transforms, such as discrete cosine transform (DCT), instead of random matrices in the weights of a neural network. Particularly, the focus relies on SSFN architecture which uses a combination of random matrices and layer-wise training to guarantee a monotonically decreasing training cost as the number of layers increases. Two methods are used to find the best deterministic transform in each layer of the network: (1) features standard deviation and (2) singular values of the correlation matrix. It is shown that both methods provide similar performance to the case of random matrices over several benchmark classification datasets. The use of deterministic transform reduces the computational complexity of matrix multiplication in each layer of the network and has been explored in other neural network architectures as well, such as scattering networks \cite{ScatteringNet_2013} and transformer encoder architecture \cite{fnet}. Further, the use of deterministic transforms allows limited learning of few parameters in the neural network, makes it suitable for a data-limited scenario.

\section{Preliminaries}    

\subsection{Deterministic Transforms}\label{subsec:Deterministic_Transforms} 
In this manuscript, the term \textit{deterministic transforms} (DT) is used to refer to discrete linear transforms used signal processing tasks. Only real transforms in one dimension are considered, such as discrete cosine transform (DCT) and discrete wavelet transform (DWT). They are expressed as linear functions and the following matrix notations are used
\begin{eqnarray}
    \mathbf{y}= \mathbf{W}_{DT}\mathbf{x} \triangleq w_{DT}(\mathbf{x})
\end{eqnarray}
where $\mathbf{W}_{DT}\in\mathbb{R}^{N\times N}$, $w_{DT}(\cdot)\in\mathbb{R}^N \rightarrow \mathbb{R}^{N}$, and $\mathbf{x} \in \mathbb{R}^N$ and $\mathbf{y} \in \mathbb{R}^N$ are respectively the input and output signal. It is worth noting that the main advantage of using deterministic transforms is a reduction in computational complexity. While a matrix-vector multiplication has a computational cost of $\mathcal{O}(N^2)$, it is possible to be reduced to $\mathcal{O}(Nlog_2N)$ \cite{wang_2012, orthogonaltransforms_DSP}, or even $\mathcal{O}(N)$ \cite{waveletT_1,WaveletT_2}, by efficiently implementing each specific deterministic transforms.

\begin{figure*}[t!]
	\centering
	\def\svgwidth{\linewidth}
	\includegraphics[width=1\textwidth]{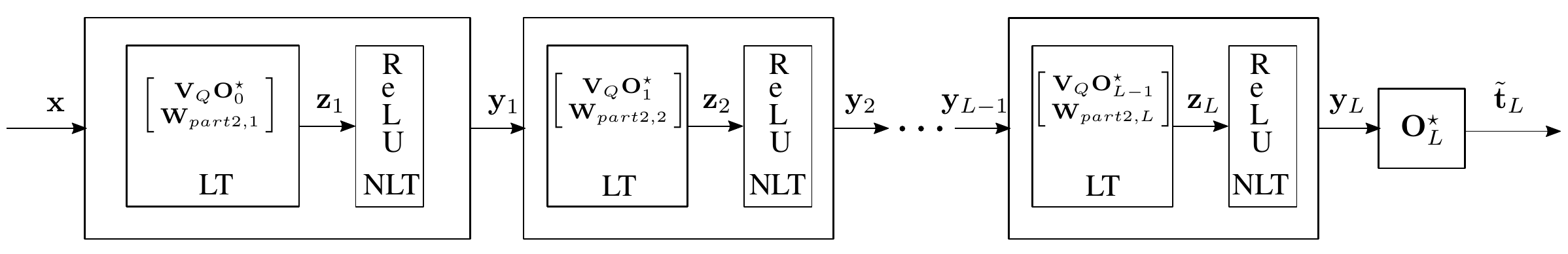}	
	\caption{The architecture of a multi-layer SSFN with $L$ layers and its signal flow diagram. LT stands for \emph{linear transform}, and NLT stands for \emph{non-linear transform} (activation function). ReLu is used as activation function.}
	\label{fig:MultiLayerPLN}
\end{figure*}
\subsection{Self Size-estimating Feed-Forward Network}
\label{subsec:SSFN} 
In this Section the training of SSFN architecture, shown in Figure~\ref{fig:MultiLayerPLN}, is quickly reviewed. Consider the signal flow between $l$'th layer and the network input as
\begin{equation}
    \mathbf{y}_l = \mathbf{g}(\mathbf{W}_{l}\mathbf{y}_{l-1}) = \mathbf{g}(\mathbf{W}_{l}...\mathbf{g}(\mathbf{W}_{2}\mathbf{g}(\mathbf{W}_{1}\mathbf{x}))...)\in\mathbb{R}^{n_l},
    \label{eq:SSFN_SignalFlow}
\end{equation}
where $\mathbf{W}_l \in \mathbb{R}^{n_l \times n_{l-1}}$ is the weight matrix of $l$'th layer with $n_l$ hidden neurons and $\mathbf{g}(\cdot)$ is ReLU activation function. The network is built using a layer-wise approach with convex optimization along with the use of random matrix instances to ensure a monotonically decreasing training cost. A new layer is added on top of the previously optimized structure and is optimized as follows:

\begin{enumerate}
    \item Consider the training dataset $\mathcal{D}=\{(\mathbf{x}^{(j)}\in\mathbb{R}^{P},\mathbf{t}^{(j)}\in\mathbb{R}^{Q})\}_{j=1}^{J}$. The $l$'th layer output is computed as in \eqref{eq:SSFN_SignalFlow}.
    \item The output matrix $\mathbf{O}_{l}$ is computed by using alternating-direction-method-of-multipliers (ADMM) by solving the following optimization problem
    \begin{equation}
        \mathbf{O}_l^\star = \text{arg}\min_\mathbf{O}\mathcal{C}_l \quad\text{s.t.} \,\,||\mathbf{O}||_F^2\leq\epsilon_l,
    \end{equation}
    where $\mathcal{C}_l=\frac{1}{J}\sum_{j=1}^J||\mathbf{t}^{(j)}-\tilde{\mathbf{t}}_l^{(j)}||^2$ defines the training cost and $\tilde{\mathbf{t}}_l^{(j)}=\mathbf{O}_l\mathbf{y}_l^{(j)}$ the prediction of $j$'th sample. Here, $\epsilon_l =  2 \alpha Q$ denotes the regularization parameter with $1 \leq \alpha$. Refer to \cite{SSFN_Saikat} for more details.
    \item The weight matrix of $(l+1)$'th layer is constructed as
        \begin{eqnarray}
            \mathbf{W}_{l+1} = 
            \left[ \begin{array}{c}
             \mathbf{V}_Q \mathbf{O}_{l}^{\star} \\
             \mathbf{R}_{l+1}
            \end{array} \right] = 
            \left[ \begin{array}{c}
             \mathbf{V}_Q \mathbf{O}_{l}^{\star} \\
             \mathbf{W}_{part2,l+1}
            \end{array} \right] ,
            \label{eq:SSFN_LinearTransform}
        \end{eqnarray}
    where $\mathbf{V}_{Q}=\left[\mathbf{I}_Q \,-\mathbf{I}_Q \right]^T \in \mathbb{R}^{2Q\times Q}$ and $\mathbf{R}_{l+1}$ refers to the computed random matrix instance. To ease generalization, the bottom part will be referred as $\mathbf{W}_{part2,l+1}$.
\end{enumerate}
This procedure is carried until the maximum number of layers $L_{max}$ is reached or the cost shows a saturation trend, $\frac{\mathcal{C}_l^*-\mathcal{C}_{l-1}^*}{\mathcal{C}_{l-1}^*}<\eta_{layer}$ with $\eta_{layer}$ being a predefined threshold. Note that this layer-wise approach avoids dealing with problems such vanishing gradients or local minima while ensuring a reduction of the cost. In \cite{SSFN_Saikat} it is shown that due to the use of $\mathbf{V}_Q$ matrix and lossless flow property (LFP) of ReLU, the training of SSFN leads to $\mathcal{C}_l \leq \mathcal{C}_{l-1}$.

\section{Proposed Learning Scheme}\label{sec:Network_Design/architecture}
Note that SSFN guarantees monotonic reduction of the training cost for any choice of the matrix $\mathbf{W}_{part2,l+1}$, whether being a random instance or not. In the article is proposed to replace the random matrix in equation \eqref{eq:SSFN_LinearTransform} with a suitable deterministic transform. Thereby it is still possible to use the advantages of LFP of ReLU activation while incorporating the low computational complexity benefits of deterministic transforms. Lets denote the weight matrix~\eqref{eq:SSFN_LinearTransform} as a linear transform block 
\begin{eqnarray}
    \mathbf{LT}_l(\cdot) = 
    \left[ \begin{array}{c}
         \mathbf{V}_Q \mathbf{O}_{l-1}^{\star} \\
         w_{DT,l}(\cdot)
    \end{array} \right],
    \label{eq:LTS_tructure}
\end{eqnarray}
where $w_{DT,l}(\cdot)$ represents a deterministic transform in $l$'th layer. Note that $\mathbf{LT}_l(\cdot)$ and $\mathbf{w}_{DT,l}(\cdot)$ are represented in function notation instead of matrix, to emphasize that is not necessarily implemented as a matrix multiplication, with $\mathcal{O}(N^2)$ complexity. 

To differentiate among the upper and lower parts of the linear transform output, it is expressed as
\begin{eqnarray}
	\mathbf{z}_{l} =
	\left[\begin{array}{c}
	    \mathbf{z}_{part1,l} \\
	    \mathbf{z}_{part2,l}
	\end{array}\right] =
	\left[\begin{array}{c}
	    \mathbf{V}_Q \mathbf{O}_{l-1}^{\star} \mathbf{y}_{l-1} \\
	    \mathbf{z}_{part2,l}
	\end{array}\right] ,
	\label{eq:Zsignal_ouputLT}
\end{eqnarray}
and the layer output $\mathbf{y}_l=\mathbf{g}(\mathbf{z}_l)$.

\begin{myremark}
    The use of ReLU as activation function $\mathbf{g}$ is a necessary condition to preserve LFP, albeit it is only necessary to be applied in the upper part $\mathbf{z}_{part1,l}$. Therefore, it is possible to use any other function as activation for $\mathbf{z}_{part2,l}$.  
\end{myremark}

Following the SSFN training, first, the layer output matrix $\mathbf{O}_l^\star$ is learned by alternating direction method of multipliers (ADMM) and $\mathbf{z}_{part1,l}$ is obtained. The second part of the linear transform $\mathbf{z}_{part2,l}$ is computed following an unsupervised approach, by the proposed Algorithm~\ref{alg:Score_system}. It defines the learning procedure to choose the most suitable transform to be used at each layer, among all other deterministic transforms available in a predefined bag formed by $T$ different transforms. 

\begin{myremark}
    Many deterministic transforms e.g., discrete cosine transform, discrete sine transform, Hartley transform, among others are squared matrices. Furthermore, some as Haar transform and Walsh-Hadamard transform are square transforms and its associated dimensions are integers of power of $2$. The transform concatenation in equation \eqref{eq:LTS_tructure} makes the network width (number of hidden neurons) monotonically increasing as the network gets deeper. To cope with this issue, node pruning based on node variance $\sigma_n$ is necessary. The variance of a single node $n$ is defined as,
    \begin{equation} \label{eq:std}
        \sigma_{n}^2 = \frac{1}{J}\sum_{j=1}^J (z_n^{(j)}-\frac{1}{J}\sum_{k=1}^J z_n^{(k)})^2, \,\,\,\, \,\, n=1,2,...,N_l,
    \end{equation}
    with $\mathbf{z}^{(j)}=\left[z_1^{(j)},...,z_n^{(j)},...,z_{N_l}^{(j)}\right]=\mathbf{z}_{part2,l}^{(j)}$. 
\end{myremark}
For each of the transforms $w_{DT,l}(\cdot)$ within the bag, the output is calculated as
\begin{equation} \label{eq:Z_part2}
    \mathbf{z}_{part2,l} = \frac{p(w_{DT,l}(\mathbf{y}_{l-1}))}{||p(w_{DT,l}(\mathbf{y}_{l-1}))||},
\end{equation}
where $p(\cdot)$ is defined as a pruning function that remove the nodes presenting low variance $\sigma^2_n<\eta_{var}$, with $\eta_{var}$ being a variance predefined threshold and $\sigma_n^2$ as defined in~\eqref{eq:std}. The normalization step is needed to arrest energy increase of signal flow through the successive layers. These two operations are critical to achieve a stable behavior for the training algorithm.

\begin{algorithm}[t!]
    \caption{: Unsupervised learning of deterministic transforms}\label{alg:Score_system}
    \begin{small}
    \mbox{Input: }
    \begin{algorithmic}[1]
        \STATE $\mathbf{y}_{l-1}$ \hfill (Input of $l$'th layer)
        \STATE $\eta_{var}$ \hfill (Variance threshold)
        \STATE Bag of deterministic transform: $DT_i$ with $i=1,2,...,T$ \label{op1:DTBox}
        \STATE $\gamma$ \hfill (Method 2 hyperparameter, in~\eqref{eq:method2})
    \end{algorithmic}
    \vspace{0.1cm}
    \mbox{Estimation of a suitable deterministic transform: }
    \vspace{-0.3cm}
    \begin{algorithmic}[1]
        \FOR{$i=1:T$}
        \STATE $w_{DT,i}=DT_i$ \hfill (Choose $i$'th DT in the bag)
        \STATE Compute $\mathbf{z}_{part2,l}$ according to \eqref{eq:Z_part2}
        \STATE Apply Method1~(\ref{subsec:Method1}) or Method2~(\ref{subsec:Method2})
        \ENDFOR
        \STATE $w_{DT,l} \leftarrow \argmin_{DT_i}(sc_1)$ \hfill (Choose DT with min $sc_1$) \label{op2:min_sc1}
    \end{algorithmic}
    \vspace{0.1cm}
    \mbox{Output: }
    \vspace{0.0cm}
    \begin{algorithmic}[1]
        \STATE $w_{DT,l}$ 
    \end{algorithmic}
    \end{small}
\end{algorithm}

Two different methods have been developed to chose, in an unsupervised way, which deterministic transform from the bag must be used at each layer. The methods assign a score $sc_1$ based on different properties of its input signal $\mathbf{z}_{part2,l}$ \eqref{eq:Z_part2}. The score is used to compare different transforms and choose a suitable one in each layer following the decision criterion in step~\ref{op2:min_sc1} in Algorithm~\ref{alg:Score_system}.
    
\subsection{Method 1: Standard deviations}\label{subsec:Method1}
Variability in layer nodes give an insight about the amount of information each node handles \cite{variance_prunning}. Nodes standard deviation $\sigma_n$ is computed by square root of equation~\eqref{eq:std}. Then standard deviation over all nodes $\sigma_T$ is computed and the score is set as
\begin{equation}
    sc_1 = \sigma_T
    \label{eq:method1}
\end{equation}
When choosing the minimum $sc_1$ among all the transforms, the deterministic transform is chosen whose information is distributed more evenly among all the nodes. One must note that nodes carrying a small amount of information do not exist since low variance nodes have been removed previously in \eqref{eq:Z_part2}. 
    
\subsection{Method 2: Singular values of cross-correlation matrix} \label{subsec:Method2}

\begin{figure}[t]
    \begin {center}
        \includegraphics[width=0.5\textwidth]{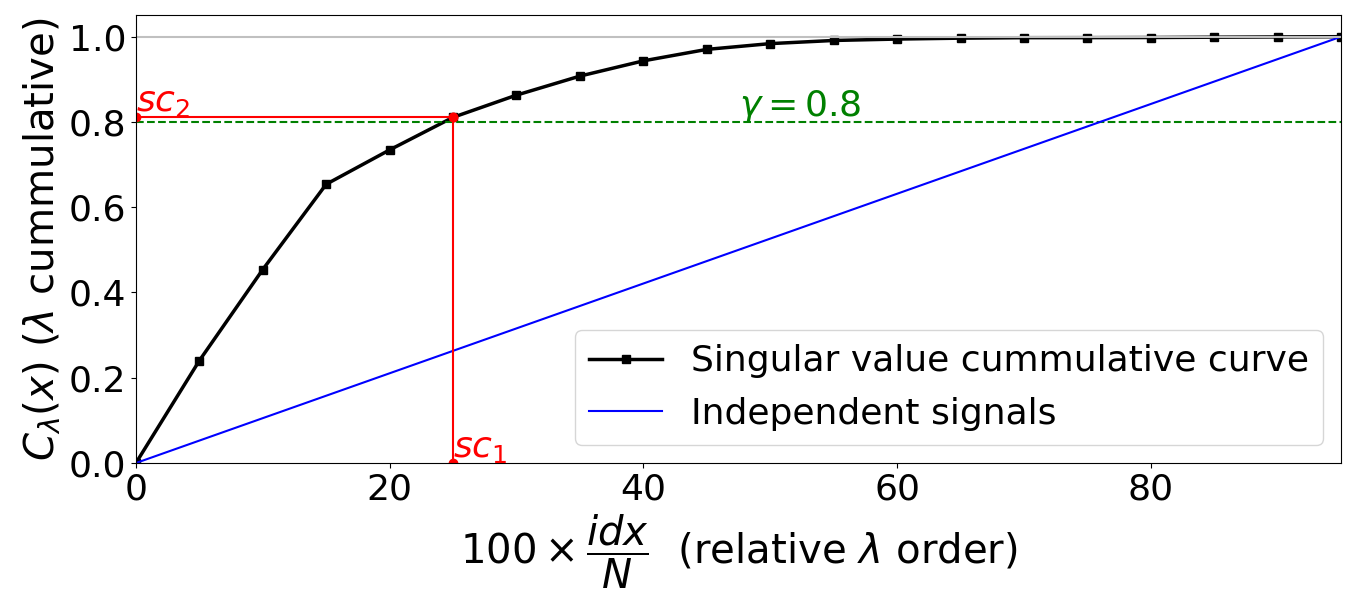}
        \vspace{-8mm}
        \caption{Example of correlation eigenvalue curve in black. In red, the first cumulative value higher than the threshold (in green) and its designated scores. In blue, a line corresponding to if the signals were uncorrelated.}
        \label{fig:method2_graphically}
    \end {center}
    \vspace{-5mm}
\end{figure}

In order to measure the information shared between network input $\mathbf{x}$ and linear transform output $\mathbf{z}_{part2,l}$, ``network input-layer output'' correlation matrix is computed as $\mathbf{R}_{\mathbf{x},\mathbf{z}} = \frac{\mathbf{C}_{\mathbf{x},\mathbf{z}}}{\sqrt{\mathbf{C}_{\mathbf{x},\mathbf{x}}*\mathbf{C}_{\mathbf{z},\mathbf{z}}}}$ where $\mathbf{C}_{\mathbf{x},\mathbf{z}}$ denotes the covariance matrix between $\mathbf{x}$ and $\mathbf{z}$, denoting $\mathbf{z}_{part2,l}=\mathbf{z}$ for simplicity. Principal component analysis (PCA) is then implemented to the correlation matrix $\mathbf{R}_{\mathbf{x},\mathbf{z}}$, obtaining the corresponding singular values $\lambda_k$, $1\leq k \leq K$, with $K$ being the total number of singular values and dimension of $\mathbf{R}_{\mathbf{x},\mathbf{z}}$. The singular values $\lambda_k$ are sorted in descend order to define the cumulative singular value as
\begin{equation}
    C_{\lambda}(k) = \frac{\sum_{i=1}^k\lambda_i}{\sum_{i=1}^K\lambda_i} \qquad \text{where}\,\,0\leq C_{\lambda}(k)\leq1.
\end{equation}
The interest relies on the deterministic transform that presents a greater value of cumulative singular values in fewer components, meaning that both signals share a greater amount of information. Score $sc_1$ is defined as the minimum index $idx$ for which the information is higher than a threshold $0\leq\gamma\leq1$, acting as a hyperparameter. It may be the case that different transforms share the same index and therefore $sc_1$. To cope with this problem a second $sc_2$ is introduced, choosing the one that shares the highest amount of information within these coefficients.
\begin{align}
\begin{split}
    &idx = \argmin_k(C_{\lambda}(k) \geq \gamma), \\
    &sc_1 = 100 \times \frac{idx}{K}, \,\,\, \mathrm{and} \,\, sc_2 = C_{\lambda}(idx),
\end{split}
\label{eq:method2}
\end{align}
where $K$ is used to avoid influence of different signal lengths presented by the different transforms output. In case of tie, the decision criterion is to chose the transform is defined as
\begin{equation}
    w_{DT,l} \leftarrow \min(sc_1), \, \max(sc_2).
\end{equation}
A visual representation of both scores is shown in Figure~\ref{fig:method2_graphically}.


\begin{table*}[t!]
    \centering
    \caption{Classification accuracy of SSFN across 50 Monte-Carlo simulations and complexity comparison between random instance and deterministic transform executions. Parameters set with minimum manual effort: $k_{max}=100$, $\alpha=2$, $\eta_{layer}=0.1$, $\eta_{var}=10^{-7}$, $L_{max}=20$, a bag of $T=11$ different transforms and Method2 hyperparameter $\gamma=0.8$.}
    \vspace{-2mm}
    \label{table:Classification_performance}
	\setlength{\tabcolsep}{7.5pt}
	\renewcommand{\arraystretch}{1.4}
	\begin{footnotesize}
	\begin{tabular}[h]{|c|c|c|c|c|c|c|c|c|}
	    \hline
	    \multirow{2}{*}{Dataset} & \multicolumn{2}{|c|}{Rand SSFN} & \multicolumn{6}{|c|}{Deterministic Transform SSFN} \\ \cline{2-9} 
	    & \multicolumn{2}{|c|}{Accuracy (in \%) (avg. $\pm$ std. dev)} & \multicolumn{4}{|c|}{Accuracy (in \%) (avg. $\pm$ std. dev)} & \multicolumn{2}{|c|}{Parameters to set} \\ \cline{2-9}
	    & Train dataset & Test dataset & \multicolumn{2}{|c|}{Method1} & \multicolumn{2}{|c|}{Method2} & $\lambda_0$ & $\mu$ \\ \cline{4-7} 
	    & & & Train dataset & Test dataset & Train dataset & Test dataset &  &  \\ \hline\hline
	    Vowel & 100 $\pm$ 0 & 60.2 $\pm$ 2.4 & 99.62 & 64.72 & 93.75 & 63.42 & $10^1$ & $10^3$ \\ \hline
	    Satimage & 95.55 $\pm$ 0.15  & 89.9 $\pm$ 0.5 & 91.41 & 89.15 & 93.62 & 89.15 & $10^6$ & $10^{8}$ \\ \cline{1-9}
        Caltech101 & 99.51 $\pm$ 0.06 & 76.1 $\pm$ 0.8 & 99.95 $\pm$ 0.02 & 76.73 $\pm$ 0.82 & 99.93 $\pm$ 0.02 & 76.39 $\pm$ 0.67 & 3 & $10^{-2}$ \\ \cline{1-9}
        Letter & 99.02 $\pm$ 0.07 & 95.7 $\pm$ 0.2 & 100 $\pm$ 0 & 92.72 $\pm$ 0.3 & 95.43 $\pm$ 0.12 & 91.07 $\pm$ 0.42 & $10^{-5}$ & $10^9$ \\ \cline{1-9}
        NORB & 99.11 $\pm$ 0.04 & 86.1 $\pm$ 0.2 & 98.22 & 84.75 & 100 & 87.46 & $10^2$ & $10^4$ \\ \cline{1-9}
        Shuttle & 99.73 $\pm$ 0.08 & 99.8 $\pm$ 0.1 & 99.8 & 99.76 & 99.96 & 99.84 & $10^5$ & $10^7$ \\ \cline{1-9}
        MNIST & 97.21 $\pm$ 0.03 & 95.7 $\pm$ 0.1 & 97.32 & 96.54 & 98.09 & 96.9 & $1$ & $10^4$ 
        \\ \hline
	\end{tabular}
	\end{footnotesize}
\end{table*}

\section{Experimental Evaluations}\label{sec:Experimental_evaluation} 
Carried experiments will compare the performance of the proposed methods to SSFN using random weights in \cite{SSFN_Saikat}, on seven different classification datasets. These datasets are chosen due to their popularity in literature and level of complexity for tasks. Vowel dataset belongs to speech recognition while the others belong to image classification. Vowel dataset has a highly limited training data size (a data-limited training scenario). Train and test dataset partitions are created using random sampling for Caltech101 and Letter datasets. For the remaining five datasets, train and test dataset partitions are already predefined. 

A bag of $T=11$ different deterministic transforms is used to train the network, $DT_i$ with $i=1,..,T$, all of them widely used in signal, image and audio processing, filtering, signal coding, among other applications. These transforms are listed below with its abbreviations and computational complexity when computed by its fast algorithms \cite{wang_2012}:
\begin{itemize}
    \item Discrete cosine and sine transforms  (DCT-II and DST-I): complexity $\mathcal{O}(Nlog_2N)$.
    \item Fast Walsh-Hadamard Transform (FWHT1 indicates the coefficients are in normal Hadamard order while FWHT2 they are in order of increasing sequency value): complexity $\mathcal{O}(Nlog_2N)$.
    \item Discrete Hartley Transform (DHT): $\mathcal{O}(Nlog_2N)$.
    \item Discrete Haar transform (Haar): complexity $\mathcal{O}(N)$.
    \item Different wavelet transforms (Daubechies 4 (DB4) and 20 (DB20), Symlets 2 (sym2), Coifflets 1 (coif1), Biorthogonal 1.3 (bior1.3) and Reverse Biorthogonal 1.1 (rbior1.1)): Complexity $\mathcal{O}(N)$.
\end{itemize}
Note that the decomposition level in wavelet transforms has been set regardless of the boundary effects, equal to $\floor{(log_2(N))}$, where $N$ is the length of the input signal or previous layer nodes. For wavelet outputs, all the obtained coefficients, approximation and detail, are concatenated.

When using the second method proposed there is an extra hyperparameter that needs to be set in~\eqref{eq:method2}. It has been set equally for all datasets as $\gamma=0.8$.

\subsection{Experimental results}
First, both methods' performances are compared against that achieved by SSFN with random weights in \cite{SSFN_Saikat} (called Rand SSFN here). Parameter tuning and results in terms of accuracy are reported in Table~\ref{table:Classification_performance}. Minimal effort has been put into tuning the parameters and only two have been carefully tuned. The parameter $\lambda_0$ is used to control the regularized LS used to obtain the output matrix in the first layer $\mathbf{O}_0^*$. The choice of $\mu$ influences the convergence of ADMM to optimize the output matrices of the successive multi-layer structure. Both of them are tuned carefully by a combination of cross-validation and manual effort. The rest of the hyperparameters of the network are states in the caption of Table \ref{table:Classification_performance}.

\begin{figure}[t]
    \begin {center}
        \includegraphics[width=0.5\textwidth]{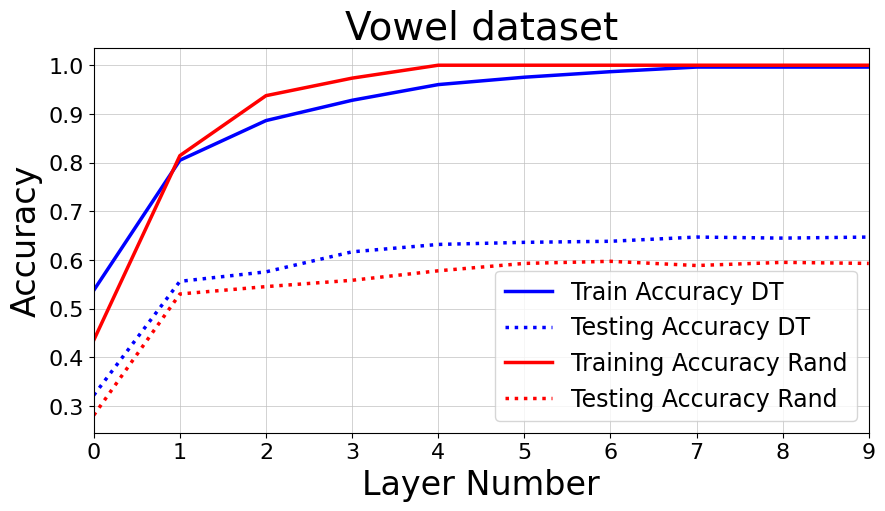}
        \vspace{-8mm}
        \caption{Training and testing accuracy curves of vowel dataset with $\eta{layer}=0.05$. In red using random instances as in~\cite{SSFN_Saikat} and in blue using deterministic transforms.}
        \label{fig:curves_acc_vowel}
    \end{center} \vspace{-5mm}
\end{figure} 

The performances presented in Table~\ref{table:Classification_performance}, both in training and testing, are close to the presented by SSFN with random weights. One must note that the standard deviation reported for the accuracy of Letter and Caltech101 datasets is due to random picking of train and test set in different Monte-Carlo experiments. For the other five datasets, train and test sets are predefined, and therefore, there is no randomness observed when using deterministic transforms during training. As an example, it is also compared the performance of SSFN throughout the layers in Figure~\ref{fig:curves_acc_vowel} for the Vowel dataset. Similar behaviour and consistency it is achieved in both cases, proving the effectiveness of the proposed approach.

In Table~\ref{table:Network_architecture} the constructed network architectures built by using \textit{Method 2} are shown. Similar results are achieved by \textit{Method 1} as well. Table~\ref{table:Network_architecture} shows the number of nodes per layer along with the deterministic transforms chosen in each layer and their corresponding test accuracy for different Monte-Carlo trials. For Vowel and MNIST datasets, the accuracy and architecture do not change in different trials due to having predefined train sets. For Letter dataset, even though the number of layers varies among executions, it does not affect the chosen transform, being DB20 in all cases. For Caltech101 dataset, the deterministic transforms chosen across executions are consistent, only differ the type of ordering for FWHT. Also, the number of layers remains equal for all the executions, presenting all of them a similar node structure.

Finally, properly tuning the parameters, as variance threshold $\eta_{var}$, can provide a further improvement in computational cost and perhaps in performance. By hand-tuning the threshold variance $\eta_{var}=10^{-6}$ for Caltech101 dataset, the network new accuracy is $76.25\pm0,84$, being `392-504-656-1125 (DB20-DB20-DCT-FWHT2)', where the number of neurons has been reduced considerably. To illustrate how the computational cost is reduced, consider an input dimension to $l$'th layer as $N_l=2^8=256$ nodes. The computational cost by the normal matrix-vector product, considering the matrix squared, is $N_l^2=65.546$ operations, while with fast algorithms is $N_llog_2N_l=2.048$ and $N_l=256$. The computations reduction is so significant that even with input nodes $N_l=2^{12}$, fast algorithms are less expensive.


\begin{table}[t!]
    \centering
	\caption{Monte-Carlo trials of SSFN with deterministic transforms applying Method2.}
	\vspace{-2mm}
	\label{table:Network_architecture}
	\setlength{\tabcolsep}{3pt}
    \renewcommand{\arraystretch}{1.2}
    \begin{footnotesize}
    \begin{tabular}{|c|c|c|}
        \hline
        \multirow{2}{*}{Dataset} & Nodes arrangement & \multirow{2}{*}{Accuracy} \\ 
        & (Deterministic Transform layout) & \\\hline\hline
        \multirow{2}{*}{Vowel} & 136-355-370-332-324 & \multirow{2}{*}{63.42} \\
        & (DB20-DB20-DB20-DB20-DB20) & \\\hline        \multirow{6}{*}{Caltech101} & 4096-558-1108-2170 & \multirow{2}{*}{76.4} \\ 
        & (FWHT2-DB20-FWHT1-FWHT2) & \\ \cdashline{2-3}
        & 4096-557-1110-2162 & \multirow{2}{*}{76.5} \\
        & (FWHT2-DB20-FWHT2-FWHT1) &\\ \cdashline{2-3}
        &  4096-556-1107-2157 & \multirow{2}{*}{76.16} \\
        & (FWHT2-DB20-FWHT2-FWHT1) &\\\hline 
        \multirow{6}{*}{Letter} & 224-454-441-451-458-464 & \multirow{2}{*}{91.78} \\
        & (DB20-DB20-DB20-DB20-DB20-DB20) & \\ \cdashline{2-3}
        & 224-454-441-452-457-469 & \multirow{2}{*}{91.11}\\
        & (DB20-DB20-DB20-DB20-DB20-DB20) &\\ \cdashline{2-3}
        & 224-454-442-454-455-464 & \multirow{2}{*}{91.39}\\
        & (DB20-DB20-DB20-DB20-DB20-DB20) &\\\hline
        \multirow{2}{*}{MNIST} & 1044-2038-1998-1931-1789 & \multirow{2}{*}{96.9} \\ 
        & ((FWHT1-FWHT2-DST-DST-FWHT2) & \\\hline 
    \end{tabular}
    \end{footnotesize}
\end{table}
    
\section{Conclusions} 
It has been shown that it is possible to employ deterministic transforms in the weight matrices of a multilayer neural network and achieve competitive classification performance on different datasets. Criterion such as features standards deviations and singular values of cross-correlation matrix between the input and output of the transform provides useful information about the power of deterministic transforms in a neural network. In this way, a new learning approach can be achieved as a hybrid combination of supervised and unsupervised learning in each layer of the network. The use of deterministic transforms reduces the computational complexity of matrix multiplication at the time of testing, making it suitable for applications with real-time or low-latency requirements. Besides, the use of deterministic transforms may provide an understanding of interpretability/explainability, and bring new insights about the information flow within layers of a neural network.

\bibliographystyle{IEEEtran}
\bibliography{references_Pol}



\end{document}